\documentclass[conference]{IEEEtran}
\IEEEoverridecommandlockouts
\usepackage{cite}
\usepackage{amsmath,amssymb,amsfonts}
\usepackage{algorithmic}
\usepackage{graphicx}
\usepackage{textcomp}
\usepackage{xcolor}
\def\BibTeX{{\rm B\kern-.05em{\sc i\kern-.025em b}\kern-.08em
    T\kern-.1667em\lower.7ex\hbox{E}\kern-.125emX}}

\usepackage{booktabs}
\usepackage{adjustbox}
\usepackage{pgfplots,pgfplotstable}
\usepgfplotslibrary{fillbetween,groupplots,colorbrewer}
\pgfplotsset{compat=1.16}
\usepackage{subcaption}

\begin{document}

\title{MAPLE: Multi-State Aggregated Policy Evaluation for AlphaZero in Imperfect-Information Games
}

\IEEEoverridecommandlockouts

\IEEEpubid{\makebox[\columnwidth]{ * 979-8-3315-9476-3/26/\textdollar31.00 \copyright2026 IEEE\hfill}\hspace{\columnsep}\makebox[\columnwidth]{ }}


\author{
  \textbf{Qian-Rong Li\textsuperscript{1,2}},
  \textbf{Hung Guei\textsuperscript{2}},
  \textbf{I-Chen Wu\textsuperscript{1}},
  \textbf{Ti-Rong Wu\textsuperscript{2}}\\
  \vspace{1pt}\\
  \textsuperscript{1}Department of Computer Science, National Yang Ming Chiao Tung University, Taiwan \\
  \textsuperscript{2}Institute of Information Science, Academia Sinica, Taiwan
  \vspace{1pt}\\
  \texttt{raschl1016.cs13@nycu.edu.tw, tirongwu@iis.sinica.edu.tw} (correspondence)\\
}

\maketitle
\IEEEpubidadjcol

\begin{abstract}
Imperfect-information games (IIGs) are challenging, as players must make decisions without fully observing the true game state.
While AlphaZero has achieved remarkable success in perfect-information games, extending it to IIGs remains difficult.
Existing search-based approaches, such as Perfect Information Monte Carlo (PIMC), suffer from strategy fusion, while Information Set Monte Carlo Tree Search (IS-MCTS) incurs high computational cost when combined with neural networks.
In this paper, we propose \textit{Multi-State Aggregated PoLicy Evaluation (MAPLE)}, a tree search method that aggregates policy and value evaluations from multiple sampled world states within a single search tree, combining the advantages of PIMC and IS-MCTS while maintaining a controllable computational cost.
We further incorporate a Siamese-based sampling strategy to select informative world states from the information set.
Experiments on Phantom Go and Dark Hex show that MAPLE significantly outperforms the PIMC-based AlphaZero baseline, achieving Elo improvements of 291 and 136, respectively.
These results demonstrate that MAPLE is an effective approach for AlphaZero-style learning in imperfect-information games.
\end{abstract}


\section{Introduction}
\label{intro}
Imperfect-information games (IIGs) are a class of challenging games in which players can only observe partial information about the game state.
Two main research directions have been explored to address IIGs.
One line of work focuses on equilibrium-based methods, such as computing Nash equilibrium using counterfactual regret minimization (CFR)~\cite{zinkevich_regret_2007,bowling_headsup_2015,brown_libratus_2017}, which have achieved strong performance in poker domains.
Although successful in several domains, these approaches are often computationally expensive.

Another direction focuses on search-based methods that reason over sampled determinizations of hidden states, including \textit{Perfect Information Monte Carlo (PIMC)}~\cite{ginsberg_gib_2001,long_understanding_2010} and \textit{Information Set Monte Carlo Tree Search (IS-MCTS)}~\cite{cowling_information_2012}.
These approaches are particularly attractive for imperfect-information board games because AlphaZero~\cite{silver_general_2018} has achieved remarkable success by combining Monte Carlo Tree Search (MCTS)~\cite{kocsis_bandit_2006,coulom_efficient_2007} with deep neural networks in perfect-information games such as Go and Chess.
Previous work, such as AlphaZe**~\cite{bluml_alphaze_2023}, has extended AlphaZero to imperfect-information board games by incorporating PIMC-based search.
While their results show that AlphaZero-style learning can be applied to imperfect-information board games, PIMC-based methods are known to suffer from issues such as strategy fusion~\cite{frank_search_1998}.
Although IS-MCTS mitigates this issue by maintaining a single search tree over the information set, directly integrating it with AlphaZero incurs a high computational cost when using neural network evaluations during the search.

To address these challenges, we propose \textit{Multi-State Aggregated PoLicy Evaluation (MAPLE)}, a search method designed to integrate AlphaZero-style learning with IIGs.
MAPLE aggregates policy and value evaluations from multiple sampled world states within a single search tree, combining the advantages of PIMC and IS-MCTS while maintaining a controllable number of network evaluations.
In addition, we introduce a Siamese-based sampling strategy, inspired by prior work~\cite{bertram_neural_2024}, to select informative world states from the information set.

Experiments on two imperfect-information board games, Phantom Go and Dark Hex, show that MAPLE significantly outperforms the existing AlphaZero-based approach AlphaZe**.
Specifically, when evaluated against the same opponents, MAPLE achieves Elo improvements of 291 in Phantom Go (increasing the win rate from 58.96\% to 88.45\%) and 136 in Dark Hex (increasing the win rate from 67.87\% to 82.20\%) compared with AlphaZe**.
These results establish a new state-of-the-art performance for AlphaZero-style learning in imperfect-information board games and highlight MAPLE as a promising direction for advancing AI in IIGs.

\section{Imperfect-Information Games}
\label{iig}

In imperfect-information games (IIGs), each player has private information hidden from the others, meaning no player has full access to the true game state.
Planning in such environments is challenging, as decisions must be made based on partial observations and uncertain knowledge of the current situation.
To address this difficulty, one line of work focuses on theoretically grounded solutions, most notably \textit{counterfactual regret minimization} (CFR) and its variants based on the Nash equilibrium~\cite{zinkevich_regret_2007,bowling_headsup_2015,brown_libratus_2017,lanctot_monte_2009,brown_combining_2020}.
Another line of work focuses on search-based methods~\cite{li_combining_2025,bluml_alphaze_2023,wang_beliefstate_2015}, which reason over possible world states during planning.
In this work, we focus on search-based methods.
The following subsections introduce the key components of these methods in IIGs.

\subsection{Information Set}
\label{iig:info_set}
In search-based methods, planning typically requires evaluating a perfect game state.
However, the true world state, i.e., the true game state, is not directly observable in IIGs.
Instead, the player only has access to partial observations and can only infer multiple possible world states based on the observed history.
These possible world states are collectively referred to as the \textit{information set}, which represents the set of states consistent with the player's view.
The size of the information set can grow exponentially, especially in games with little information revealed, resulting in a large number of possible states to consider during search.

Handling the full information set is intractable in most IIGs.
A common approach is therefore to approximate the information set by sampling a subset of possible world states for search.
A naive strategy is to uniformly sample world states from the information set.
For example, prior work samples possible opponent moves in Phantom Go~\cite{cazenave_phantomgo_2006}, or samples card deals with prior observations in Bridge~\cite{ginsberg_gib_1999}.
However, such approaches often generate states that are far from the true world state, leading to inaccurate or inconsistent evaluations during search.

To improve sampling quality, more advanced methods assign different importance to candidate world states based on their likelihood.
For example, recent work~\cite{bertram_neural_2024} employs representation learning in Reconnaissance Blind Chess (RBC) to capture similarities between states by using a Siamese network to encode world states into an embedding space.
Specifically, an \textit{anchor} state derived from partial observations is compared with a \textit{positive} sample corresponding to the true world state and several \textit{negative} samples that are possible but inconsistent with the true state.
This training objective allows the model to learn meaningful relationships between states in the information set.
The representation helps prioritize states that are more consistent with the current observations, improving the effectiveness of sampling possible world states.

\subsection{Tree Search Methods}
\label{iig:tree_search_method}
Given a set of sampled world states from the information set, \textit{Perfect Information Monte Carlo (PIMC)}~\cite{frank_search_1998,ginsberg_gib_2001} performs search independently on each sampled world state.
Any search algorithm can be applied.
The final decision is obtained by aggregating the search results from all sampled world states, i.e., by averaging the values for each action.
PIMC has been successfully applied in several IIGs~\cite{ginsberg_gib_2001,buro_improving_2009}, however, previous research~\cite{frank_search_1998} has also identified two of its fundamental problems: \textit{strategy fusion} and \textit{non-locality}.

Strategy fusion arises because the search is conducted independently on each sampled world state, which may favor different actions across states.  
When these results are aggregated into a single decision for the information set, the resulting action can be inconsistent with the optimal choice for any individual world state.
On the other hand, non-locality occurs because PIMC does not account for the opponent's knowledge during search.

To address strategy fusion, \textit{Information Set Monte Carlo Tree Search} (IS-MCTS)~\cite{whitehouse_determinization_2011,cowling_information_2012} was proposed.
It constructs a single search tree where each node represents an information set rather than a specific determinized state.
In each iteration, a world state is sampled from the information set, and a standard MCTS iteration is performed based on this sampled world state while updating a shared tree.
Specifically, each iteration follows the four standard steps of \textit{Monte Carlo tree search} (MCTS)~\cite{kocsis_bandit_2006,coulom_efficient_2007}: \textit{selection}, \textit{expansion}, \textit{simulation}, and \textit{backpropagation}.
During selection, it traverses the tree from the root to a leaf node by selecting child nodes according to their UCT scores.
During expansion, a new child node is added to the selected leaf node.  
During simulation, the value of the new node is estimated by a rollout, i.e., playing out the game randomly to a terminal state.
During backpropagation, the rollout result is propagated back to update the statistics of all visited nodes.
The four steps above are repeated as long as the thinking budget remains.
Eventually, MCTS selects the action based on the root node statistics.
By maintaining a unified search tree, IS-MCTS reduces the inconsistency caused by independent state searches in PIMC, and has been widely applied to imperfect-information games~\cite{bitan_combining_2018,li_combining_2025}.

\subsection{Zero-Knowledge Learning Methods}
\label{iig:zero_knowledge}
AlphaZero~\cite{silver_general_2018} is a zero-knowledge learning method that provides a general framework for training agents from scratch without human knowledge, achieving or even surpassing human performance.
It improves MCTS by integrating deep learning to guide the tree search, predicting the \textit{policy} and \textit{value} of each state.
This enables AlphaZero to explore promising states more effectively, making it highly suitable for scenarios requiring deep lookahead, such as strategic board games.
During training, AlphaZero continuously runs \textit{self-play} and \textit{optimization}: self-play collects data using tree search, while optimization uses this data to update the network model.

For self-play, AlphaZero runs a modified MCTS as follows.
In selection, at each state $s$, it selects the child node based on the PUCT~\cite{silver_mastering_2017,rosin_multiarmed_2011} formula:
\begin{equation}\label{eq:puct}
\mathop{\arg\max}_{a} \bigl\{ Q(s,a) + c_\mathrm{puct}P(s,a)\sqrt{\frac{\sum_{b}{N(s,b)}}{1+N(s,a)}} \bigr\},
\end{equation}
where $Q(s,a)$, $P(s,a)$, and $N(s,a)$ are the mean value, prior knowledge from policy network, and visit count of a child node $a$; and $c_\mathrm{puct}$ is a constant.
In expansion and simulation, rather than obtaining the value with a rollout, AlphaZero uses its network to predict the policy and value for the selected leaf node.
Finally, the value is used in the backpropagation along the selection path.
For optimization, AlphaZero trains the policy and value network, where policy $p$ aims to approximate the improved policy $\pi$ obtained from MCTS, and value $v$ minimizes the error between the predicted value and the game outcome $z$.

AlphaZero has achieved superhuman performance in many perfect-information board games.
For IIGs, AlphaZe**~\cite{bluml_alphaze_2023} extends AlphaZero by adapting a modified PIMC that aggregates search results using the average policy rather than the average value.
This is because the value estimates across sampled perfect-information states are not normalized and may be biased toward favorable determinizations.
This approach demonstrates that AlphaZero-style methods can be successfully applied to IIGs and provides a simple and robust baseline for zero-knowledge learning.

\section{Multi-state Aggregated Policy Evaluation}
\label{maple}

We present \textit{Multi-state Aggregated PoLicy Evaluation (MAPLE)}, an approach extending AlphaZero to IIGs by combining the advantages of PIMC and IS-MCTS.

\subsection{Imperfect-Information Search with AlphaZero}
\label{maple:iis_az}

One key challenge when integrating imperfect-information games (IIGs) search methods with the AlphaZero framework is the resulting computational cost.
In AlphaZero, MCTS relies heavily on the policy and value networks to guide the search, and each node expansion requires a neural network evaluation.
Consequently, the computational cost is largely determined by the number of network evaluations.

We consider two widely used approaches as described in Section~\ref{iig:tree_search_method}: PIMC and IS-MCTS.
For PIMC-based methods, such as AlphaZe**, the algorithm first samples $k$ determinized world states from the information set.
For each determinized world state, an independent AlphaZero search is performed with $N$ simulations.
As a result, the total number of network evaluations is $k \times N$.

Next, we consider another approach, IS-MCTS, which samples a different determinized world state for every simulation and performs selection based on that sampled world state.
Although IS-MCTS addresses the strategy fusion issue in PIMC by maintaining a single search tree over information sets, controlling the number of network evaluations becomes challenging when integrating it with AlphaZero.
Since the policy prior, $P(s,a)$, used in the PUCT formula depends on the sampled determinization in each simulation, the neural network must be evaluated along the entire selection path.
As a result, the total number of network evaluations becomes $N \times d$, where $d$ is the average search depth.
Unlike the PIMC-based method, the value of $d$ cannot be directly controlled, as it depends on the structure of the search tree and varies across different states, making it difficult to precisely control the number of network evaluations.
In summary, PIMC provides a controllable computational cost but suffers from the strategy fusion issue, while IS-MCTS mitigates strategy fusion but makes the computational cost difficult to control.

\begin{figure*}[!ht]
    \centering
    \includegraphics[width=\linewidth]{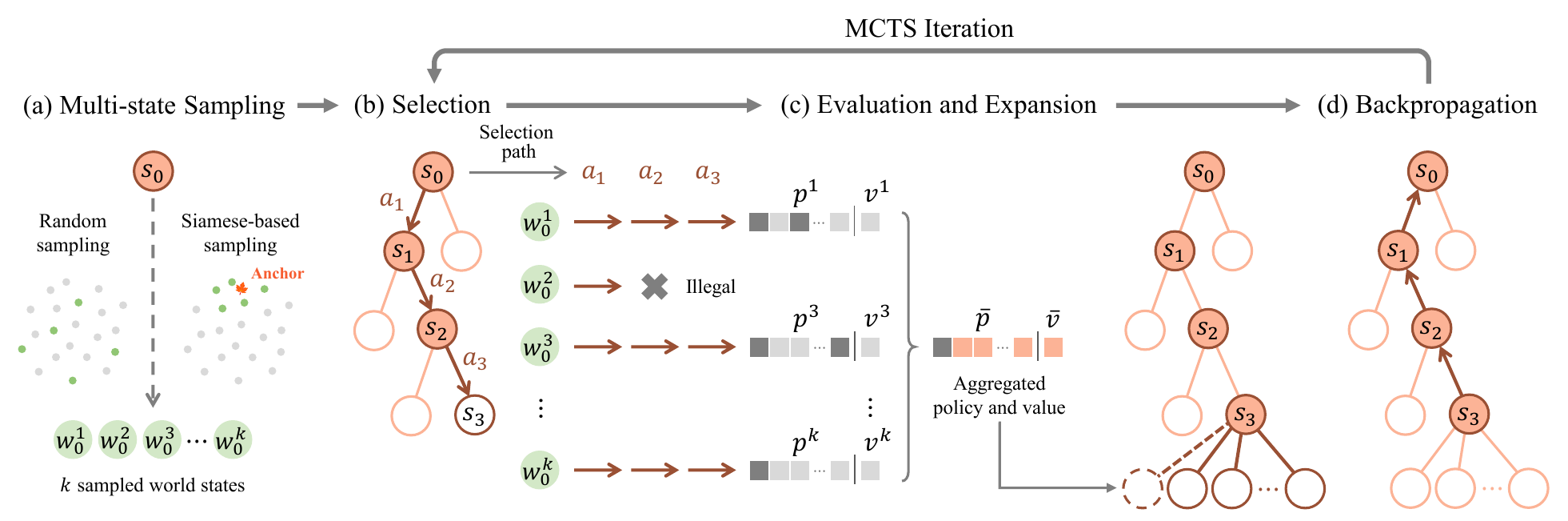}
    \caption{Overview of MAPLE tree search.
    (a) $k$ possible world states are sampled from the information set using random or Siamese-based sampling.
    (b) The selected action sequence is applied to each state, discarding illegal ones.
    (c) Valid states are evaluated by the network, and their policy and value are aggregated for expansion. Black grids indicate illegal actions.
    (d) The aggregated value is backpropagated along the selection path.}
    \label{fig:maple}
\end{figure*}

\subsection{MAPLE Tree Search}
\label{maple:tree_search}
To combine the advantages of PIMC and IS-MCTS, we propose \textit{Multi-state Aggregated PoLicy Evaluation (MAPLE)}, which is designed to provide a controllable computational cost in terms of network evaluations when integrated with the AlphaZero algorithm, while maintaining a single aggregated search tree to mitigate the strategy fusion issue.
As illustrated in Fig.~\ref{fig:maple}, the tree search consists of four phases: an initial phase for information set sampling, followed by $N$ simulations, each consisting of three phases: selection, evaluation and expansion, and backpropagation.
The following paragraphs describe each phase in detail.

\textbf{Multi-state Sampling.}
In IIGs, the true game state is unknown to both players.
Therefore, we first sample $k$ determinized world states from the information set based on the current information state $S$ to conduct the MAPLE tree search, as shown in Fig.~\ref{fig:maple}a.
Various sampling methods can be adopted for this step, as introduced in Section~\ref{iig:info_set}.
The $k$ sampled world states, denoted as $w^1, w^2, \dots, w^k$, represent possible true game states for the subsequent tree search.

\textbf{Selection.}
Similar to IS-MCTS, MAPLE aims to maintain a single search tree where each node represents an information set rather than a specific world state.
All sampled $k$ world states share this tree during the search.
In the selection phase, we first select a path (e.g., $a_1, a_2, a_3$ in Fig.~\ref{fig:maple}b) following the PUCT formula in~\eqref{eq:puct}.
Next, the selected action sequences are applied to each sampled world state $w^i$.
Since the tree is abstract and does not correspond to a specific world state, some actions along the selection path may not be legal for certain sampled states (e.g., executing $a_1, a_2$ is illegal from $w_0^2$).
Such states are discarded during the selection phase.
The remaining valid world states are used in the subsequent evaluation and expansion phase.

\textbf{Evaluation and Expansion.}
Similar to AlphaZero, MAPLE uses a policy and value network to evaluate leaf nodes.
Each valid world state $w^i$ that reaches the leaf node is evaluated by the network to obtain a policy distribution $p^i$ and value $v^i$.
Since each $w^i$ corresponds to a perfect-information environment, the set of legal actions may differ across world states.
To obtain a unified evaluation for the information set node, the policies and values from all valid world states are aggregated.
Specifically, the aggregated policy $\bar{p}(a_j)$ for each child action $a_j$ is computed as
\begin{equation}
    \bar{p}(a_j) = \frac{1}{|W_j|} \sum_{w^i \in W_j} p^i(a_j),
\end{equation}
where $W$ denotes the set of valid world states reaching the leaf node, $W_j \subseteq W$ represents the subset of valid world states in which action $a_j$ is legal, and $p^i(a_j)$ is the policy probability of action $a_j$ predicted for $w^i$.
If an action $a_j$ is illegal for all world states, the corresponding child node is not expanded.

Next, the aggregated value $\bar{v}$ is computed as
\begin{equation}  
    \bar{v} = \frac{1}{|W|} \sum_{w^i \in W} v^i,
\end{equation}
where $v^i$ is the value predicted for $w^i$.
The aggregated policy $\bar{p}$ is used to initialize the prior probabilities of the child nodes during expansion, while $\bar{v}$ is used for the subsequent backpropagation.

\textbf{Backpropagation.}
Finally, the aggregated value $\bar{v}$ is backpropagated to each node along the selection path.
The visit count and value are updated following the standard MCTS backup procedure used in AlphaZero.

In summary, MAPLE aggregates multi-state evaluation within a single search tree to mitigate strategy fusion, while maintaining a controllable computational cost by evaluating only a fixed number of sampled world states.

\subsection{Integration with AlphaZero Training}
\label{maple:intergrate_alphazero}
MAPLE can be easily integrated into the standard AlphaZero training framework.
The training process follows the self-play and network optimization pipeline.
During self-play, the agent performs a MAPLE tree search at each time step to obtain an aggregated policy at the root node.
Once a game is finished, the trajectories, together with the search policy $\pi$ and the outcome $z$, are stored in a replay buffer and used to train the policy and value network.

One key difference between MAPLE and the standard AlphaZero framework is the additional information set sampling phase before the tree search.
In this work, we consider two sampling strategies: \textit{random sampling} and \textit{Siamese-based sampling}.
For random sampling, we randomly sample $k$ distinct possible world states from the information set.
This approach does not require additional learning components and serves as a simple baseline strategy.

On the other hand, Siamese-based sampling aims to improve the quality of sampled world states.
In previous work~\cite{bertram_neural_2024}, the network was trained using a large collection of game records to learn the consistency between observed histories and possible world states.
However, for many imperfect-information games, such rich datasets may not be available.
Moreover, as the AlphaZero agent improves during training, the distribution of possible world states may also change over time.

To address these issues, we propose to train the Siamese network jointly with the AlphaZero learning process.
During the information set sampling, we first generate $m$ candidate world states from the information set.
The Siamese network computes the embedding distance between the current information state (\textit{anchor}) and each candidate state.
The $k$ candidates with the smallest distances are selected as the sampled world states for the MAPLE tree search.
Although evaluating $m$ candidate world states introduces a small additional computational cost, this filtering step improves the quality of sampled world states and leads to more consistent evaluations.

Next, the Siamese network is trained together with the policy and value networks during optimization.
For each sampled training state from the trajectory, we randomly sample a candidate state from the information set and apply the triplet loss across the imperfect-information observation (\textit{anchor}), the true game state (\textit{positive}), and the sampled candidate state (\textit{negative}).
As training progresses, the Siamese network continuously learns from the replay buffer.
This allows the sampling strategy to adapt to the evolving policy of the AlphaZero agent and select more representative world states for MAPLE search.
Overall, the loss for MAPLE is as follows:
\begin{equation}\label{eq:loss}
\mathcal{L} = (z - v)^2 - \pi^{\mathsf{T}} \log p + \max\!\left(0,\; 1 + d_{ap} - d_{an}\right) + c\|\theta\|^2,
\end{equation}
where the first three terms correspond to the value, policy, and Siamese losses, respectively.
Here, $d_{ap}$ denotes the Euclidean distance between anchor and positive embeddings, and $d_{an}$ denotes the Euclidean distance between anchor and negative embeddings.
The last term is an L2 regularization term.

\section{Experiments}
\label{experiments}

\subsection{Experimental Environments}
\label{exp:environments}
We consider two IIGs: \textit{Phantom Go} and \textit{Dark Hex}.
The rules and feature representation are described as follows.

\subsubsection{Phantom Go}
Phantom Go is an imperfect-information variant of Go.
The rules are identical to standard Go: two players alternately place stones on the board, except that players cannot see the opponent's stones.
The game is managed by a referee who observes the true board state.
When a player attempts to place a stone, the referee informs the player whether the move is \textit{legal} or \textit{illegal}.
For a legal move, the stone is placed on the board, and any captured stones are revealed to both players.
For an illegal move, the referee reports the reason, such as attempting to place a stone on an occupied position, making a \textit{suicide} move, or violating the \textit{ko} rule.
The player can then choose another position and continue attempting moves until a legal move is made.
The board size is 9$\times$9, and the \textit{komi} is set to 1 in our experiments.

\subsubsection{Dark Hex}
Dark Hex is an imperfect-information variant of Hex.
The objective of the game is to connect two opposite sides of the board with a continuous chain of stones.
The information feedback rule is similar to Phantom Go: players cannot observe the opponent's stones and only receive feedback on whether a move is legal or illegal.
If a player attempts to place a stone on an occupied position by the opponent, the move is rejected by the referee and reported as illegal.
The board size is 11$\times$11 in our experiments.

\subsubsection{Feature Representation}
For both games, we use the same feature representation for the AlphaZero network.
The input consists of six feature planes: four \textit{board planes} encoding the current board state and two \textit{revealed information planes} indicating the stones revealed by both players.
The Siamese network simply uses four board planes for \textit{positive} and \textit{negative} boards.
For the \textit{anchor} input, we use historical observations, encode the last eight turns, resulting in 34 and 18 planes for Phantom Go and Dark Hex, respectively.
For Phantom Go, we additionally record tried position (illegal moves reported by the referee) and captured positions to encode Phantom Go-specific information.
The detailed feature definitions are summarized in Table~\ref{tab:features}.

\begin{table}[t]
\centering
\caption{Features used in the experiments. Board and revealed information features are used by the AlphaZero network, while history features are used as the Siamese anchor input. Positive and negative inputs use board features.}

\begin{tabular}{lcl}
\toprule

\textbf{Feature} & \textbf{\# of planes}\\
\midrule

\textbf{Board} \\
Our Stone & 1 \\
Opponent Stone & 1 \\
Player Turn & 2 \\

\\
\textbf{Revealed information} \\
Known Opponent Stone & 1 \\
Opponent-known Stone & 1 \\

\midrule
\textbf{History observation (last 8 turns)} \\
Our Stone & 8 \\
Known Opponent Stone & 8 \\
Tried Position (Phantom Go-only) & 8 \\
Captured Stone (Phantom Go-only) & 8 \\
Player Turn & 2 \\

\bottomrule
\end{tabular}
\label{tab:features}
\end{table}

\subsection{Performance of MAPLE}
\label{exp:performance}
We implement the MAPLE tree search and the Siamese network on an open-source AlphaZero training framework~\cite{wu_minizero_2025}.
To accelerate training, we adopt the Gumbel AlphaZero algorithm~\cite{danihelka_policy_2022}, which ensures policy improvement even with a relatively small number of search simulations.
We evaluate the performance of the proposed approach by training three models: (a) \textit{AlphaZe**}, a PIMC-based AlphaZero approach serving as the baseline, (b) \textit{MAPLE-r}, which uses the MAPLE tree search with random sampling for information set sampling, and (c) \textit{MAPLE}, which uses the MAPLE tree search with Siamese-based sampling.
The number of sampled world states $k$ is set to 5 for all models, and the number of candidate world states $m$ is set to 50 for MAPLE.

Each model is trained on two IIGs introduced in Section~\ref{exp:environments}.
The policy and value network is a residual network consisting of three blocks with 256 convolutional filters.
The number of simulations $N$ is set to 16 for each move during self-play games, resulting in a total of $k \times N = 5 \times 16 = 80$ network evaluations.
For MAPLE, an additional cost of 50 Siamese network evaluations is required for each move.
The training process contains 200 iterations, where each iteration generates 1,000 self-play games and performs 200 network optimization steps.
Other hyperparameters are listed in Table~\ref{tab:hyperparameter}.

\begin{table}[t]
    \caption{Hyperparameters used in training.}
    \centering
    \begin{tabular}{lcc}
        \toprule
        Parameter & Phantom Go & Dark Hex\\
        \midrule
        \textbf{Self-Play}\\
        \# Simulations & \multicolumn{2}{c}{16}\\
        \# Games & \multicolumn{2}{c}{200K games}\\
        \# Network blocks & \multicolumn{2}{c}{3}\\
        \# Network filters & \multicolumn{2}{c}{256}\\
        \# Gumbel sampled size & 16 & 8\\
        \midrule
        \textbf{Optimization}\\
        Optimizer & \multicolumn{2}{c}{SGD}\\
        Training steps & \multicolumn{2}{c}{40K}\\
        Learning rate & \multicolumn{2}{c}{0.02}\\
        Momentum & \multicolumn{2}{c}{0.9}\\
        Weight decay & \multicolumn{2}{c}{0.0001}\\
        Batch size & \multicolumn{2}{c}{1024}\\
        Replay buffer size & \multicolumn{2}{c}{20k games}\\
        \midrule
        \textbf{MAPLE}\\
        \# candidate world states ($m$) & \multicolumn{2}{c}{50}\\
        \# sampled world states ($k$) & \multicolumn{2}{c}{5}\\
        \bottomrule
    \end{tabular}
    \label{tab:hyperparameter}
\end{table}

After training, we evaluate the playing strength of the three models.
For a fair comparison, all models are evaluated against the same pool of six opponents\footnote{Since Phantom Go converges faster, we use the checkpoints at 5k and 15k training steps. For Dark Hex, we use the checkpoints at 15k and 25k steps.}, where each of the three models contributes two checkpoints.
During evaluation, the number of simulations is set to 100 per move, and each model checkpoint plays 250 games against each opponent, with the first player and second player alternating for fairness.
As a result, each checkpoint plays 1,500 evaluation games in total.

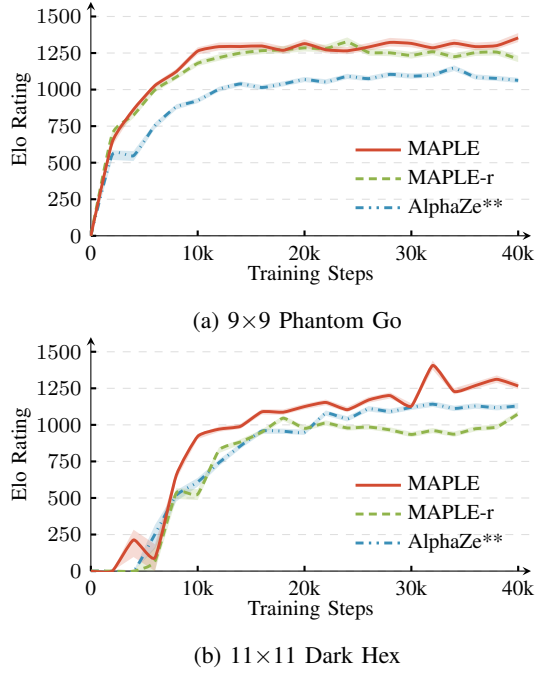
\begin{figure}[!t]
    \centering
    \begin{subfigure}[]{0.88\linewidth}
        \begin{tikzpicture}
            \begin{axis}[
                axis x line=bottom,
                axis y line=left,
                xlabel={Training Steps},
                ylabel={Elo Rating},
                xlabel style={yshift=5pt},
                ylabel style={yshift=-5pt},
                label style={font=\footnotesize},
                tick label style={font=\footnotesize},
                xmin=0, xmax=41200,
                ymin=0, ymax=1600,
                xtick={0, 10000, 20000, 30000, 40000},
                xticklabels={0,10k,20k,30k,40k},
                yticklabel style={/pgf/number format/1000 sep={}},
                scaled x ticks=false,
                legend style={
                    font=\footnotesize,
                    draw=none,
                    fill=none},
                legend pos=south east,
                legend cell align={left},
                reverse legend=true,
                ytick={0, 250, 500, 750, 1000, 1250, 1500},
                ymajorgrids=true,
                grid style={dashed, gray!25},
                width=0.95\linewidth,
                height=0.6\linewidth,
                line width=0.9pt,
                tick style={black, line width=0.8pt},
                tick align=inside,
                major tick length=2pt,
                mark size=3pt
            ]
            \definecolor{blue_az}{RGB}{58,143,183}
            \definecolor{red_maple}{RGB}{210,75,49}
            \definecolor{green_maple_r}{RGB}{144,180,75}
            
            \addplot[name path=AZ_low, blue_az, opacity=0, forget plot] coordinates {(0,0.00)(2000,518.96)(4000,511.35)(6000,731.92)(8000,863.42)(10000,907.34)(12000,985.50)(14000,1021.63)(16000,996.70)(18000,1021.52)(20000,1051.41)(22000,1035.20)(24000,1072.05)(26000,1057.19)(28000,1086.08)(30000,1074.48)(32000,1081.69)(34000,1126.94)(36000,1068.10)(38000,1058.75)(40000,1045.55)};
            \addplot[name path=AZ_high, blue_az, opacity=0, forget plot] coordinates {(0,0.00)(2000,584.59)(4000,579.47)(6000,774.90)(8000,899.48)(10000,942.17)(12000,1019.49)(14000,1056.03)(16000,1031.31)(18000,1056.09)(20000,1086.59)(22000,1070.21)(24000,1108.01)(26000,1092.53)(28000,1122.56)(30000,1110.44)(32000,1118.00)(34000,1165.47)(36000,1103.94)(38000,1094.36)(40000,1080.67)};
            \addplot[blue_az, opacity=0.2, forget plot] fill between[of=AZ_low and AZ_high];
            \addplot[color=blue_az, dashdotdotted, mark=none, smooth, tension=0.3, line width=1.1pt] coordinates {(0,0.00)(2000,554.42)(4000,548.27)(6000,754.22)(8000,881.75)(10000,924.94)(12000,1002.49)(14000,1038.74)(16000,1013.97)(18000,1038.71)(20000,1068.82)(22000,1052.57)(24000,1089.79)(26000,1074.67)(28000,1104.04)(30000,1092.22)(32000,1099.58)(34000,1145.78)(36000,1085.80)(38000,1076.35)(40000,1062.95)};
            \addlegendentry{AlphaZe**}
            
            \addplot[name path=MR_low, green_maple_r, opacity=0, forget plot] coordinates {(0,0.00)(2000,669.94)(4000,806.06)(6000,980.52)(8000,1068.57)(10000,1160.25)(12000,1197.73)(14000,1227.28)(16000,1242.48)(18000,1248.07)(20000,1260.63)(22000,1255.78)(24000,1297.21)(26000,1231.58)(28000,1228.51)(30000,1211.90)(32000,1235.65)(34000,1203.55)(36000,1229.84)(38000,1233.69)(40000,1188.04)};
            \addplot[name path=MR_high, green_maple_r, opacity=0, forget plot] coordinates {(0,0.00)(2000,718.86)(4000,844.43)(6000,1014.50)(8000,1104.30)(10000,1201.05)(12000,1241.72)(14000,1274.96)(16000,1291.96)(18000,1298.10)(20000,1312.78)(22000,1307.13)(24000,1356.78)(26000,1279.42)(28000,1276.40)(30000,1257.80)(32000,1284.11)(34000,1248.68)(36000,1277.86)(38000,1281.50)(40000,1231.16)};
            \addplot[green_maple_r, opacity=0.2, forget plot] fill between[of=MR_low and MR_high];
            \addplot[color=green_maple_r, mark=none, densely dashed, smooth, tension=0.3, line width=1.1pt] coordinates {(0,0.00)(2000,695.62)(4000,825.74)(6000,997.52)(8000,1086.21)(10000,1180.08)(12000,1218.95)(14000,1250.11)(16000,1266.08)(18000,1271.91)(20000,1285.38)(22000,1280.19)(24000,1325.13)(26000,1254.47)(28000,1251.43)(30000,1233.96)(32000,1258.81)(34000,1225.28)(36000,1252.82)(38000,1256.56)(40000,1208.88)};
            \addlegendentry{MAPLE-r}
            
            \addplot[name path=M_low, red_maple, opacity=0, forget plot] coordinates {(0,0.00)(2000,611.80)(4000,845.68)(6000,1010.76)(8000,1104.53)(10000,1239.81)(12000,1268.57)(14000,1269.26)(16000,1272.81)(18000,1245.10)(20000,1287.76)(22000,1248.27)(24000,1242.04)(26000,1266.37)(28000,1295.35)(30000,1289.32)(32000,1261.33)(34000,1290.54)(36000,1269.13)(38000,1275.62)(40000,1325.07)};
            \addplot[name path=M_high, red_maple, opacity=0, forget plot] coordinates {(0,0.00)(2000,666.25)(4000,881.95)(6000,1045.08)(8000,1141.84)(10000,1288.91)(12000,1321.95)(14000,1322.22)(16000,1326.70)(18000,1294.27)(20000,1344.05)(22000,1298.46)(24000,1290.26)(26000,1318.61)(28000,1353.62)(30000,1345.37)(32000,1313.20)(34000,1346.06)(36000,1321.16)(38000,1329.04)(40000,1386.40)};
            \addplot[red_maple, opacity=0.2, forget plot] fill between[of=M_low and M_high];
            \addplot[color=red_maple, mark=none, solid, smooth, tension=0.3, line width=1.1pt] coordinates {(0,0.00)(2000,640.68)(4000,864.17)(6000,1027.85)(8000,1122.84)(10000,1263.25)(12000,1293.85)(14000,1294.35)(16000,1298.30)(18000,1268.55)(20000,1314.27)(22000,1272.18)(24000,1265.07)(26000,1291.14)(28000,1322.71)(30000,1315.72)(32000,1285.96)(34000,1316.70)(36000,1293.80)(38000,1300.90)(40000,1353.66)};
            \addlegendentry{MAPLE}
            
            \end{axis}
        \end{tikzpicture}
        \caption{9$\times$9 Phantom Go}
        \label{fig:phantomgo_elo}
        \end{subfigure}

    \begin{subfigure}[]{0.88\linewidth}
        \begin{tikzpicture}
            \begin{axis}[
                axis x line=bottom,
                axis y line=left,
                xlabel={Training Steps},
                ylabel={Elo Rating},
                xlabel style={yshift=5pt},
                ylabel style={yshift=-5pt},
                label style={font=\footnotesize},
                tick label style={font=\footnotesize},
                xmin=0, xmax=41200,
                ymin=0, ymax=1600,
                xtick={0, 10000, 20000, 30000, 40000},
                xticklabels={0,10k,20k,30k,40k},
                yticklabel style={/pgf/number format/1000 sep={}},
                scaled x ticks=false,
                legend style={
                    font=\footnotesize,
                    draw=none,
                    fill=none},
                legend pos=south east,
                legend cell align={left},
                reverse legend=true,
                ytick={0, 250, 500, 750, 1000, 1250, 1500},
                ymajorgrids=true,
                grid style={dashed, gray!25},
                width=0.95\linewidth,
                height=0.6\linewidth,
                line width=0.9pt,
                tick style={black, line width=0.8pt},
                tick align=inside,
                major tick length=2pt,
                mark size=3pt
            ]
            \definecolor{blue_az}{RGB}{58,143,183}
            \definecolor{red_maple}{RGB}{210,75,49}
            \definecolor{green_maple_r}{RGB}{144,180,75}
            
            \addplot[name path=AZ_low, blue_az, opacity=0, forget plot] coordinates {(0,0.00)(2000,0.00)(4000,0.00)(6000,151.96)(8000,476.31)(10000,576.82)(12000,718.35)(14000,835.98)(16000,939.32)(18000,938.84)(20000,931.22)(22000,1062.28)(24000,1023.33)(26000,1091.39)(28000,1073.49)(30000,1101.55)(32000,1123.02)(34000,1093.91)(36000,1110.35)(38000,1099.00)(40000,1111.39)};
            \addplot[name path=AZ_high, blue_az, opacity=0, forget plot] coordinates {(0,0.00)(2000,0.00)(4000,0.00)(6000,316.13)(8000,552.14)(10000,637.16)(12000,763.82)(14000,874.30)(16000,974.79)(18000,974.32)(20000,966.81)(22000,1098.43)(24000,1058.79)(26000,1128.37)(28000,1109.93)(30000,1138.89)(32000,1161.22)(34000,1130.98)(36000,1148.02)(38000,1136.24)(40000,1149.10)};
            \addplot[blue_az, opacity=0.2, forget plot] fill between[of=AZ_low and AZ_high];
            \addplot[color=blue_az, dashdotdotted, smooth, tension=0.3, mark=none, line width=1.1pt] coordinates {(0,0.00)(2000,0.00)(4000,0.00)(6000,252.31)(8000,517.86)(10000,609.10)(12000,742.02)(14000,855.56)(16000,957.17)(18000,956.70)(20000,949.15)(22000,1080.14)(24000,1040.95)(26000,1109.58)(28000,1091.46)(30000,1119.89)(32000,1141.71)(34000,1112.14)(36000,1128.82)(38000,1117.29)(40000,1129.88)};
            \addlegendentry{AlphaZe**}
            
            \addplot[name path=MR_low, green_maple_r, opacity=0, forget plot] coordinates {(0,0.00)(2000,0.00)(4000,0.00)(6000,0.00)(8000,494.30)(10000,485.50)(12000,807.46)(14000,864.29)(16000,932.18)(18000,1028.02)(20000,958.63)(22000,995.39)(24000,960.97)(26000,967.98)(28000,947.83)(30000,916.28)(32000,943.11)(34000,918.22)(36000,955.82)(38000,967.51)(40000,1058.42)};
            \addplot[name path=MR_high, green_maple_r, opacity=0, forget plot] coordinates {(0,0.00)(2000,0.00)(4000,0.00)(6000,165.17)(8000,567.02)(10000,559.73)(12000,847.11)(14000,901.51)(16000,967.75)(18000,1063.53)(20000,993.92)(22000,1030.63)(24000,996.25)(26000,1003.21)(28000,983.21)(30000,952.12)(32000,978.54)(34000,954.03)(36000,991.13)(38000,1002.75)(40000,1094.48)};
            \addplot[green_maple_r, opacity=0.2, forget plot] fill between[of=MR_low and MR_high];
            \addplot[color=green_maple_r, mark=none, densely dashed, smooth, tension=0.3, line width=1.1pt] coordinates {(0,0.00)(2000,0.00)(4000,0.00)(6000,68.42)(8000,533.96)(10000,526.07)(12000,827.80)(14000,883.22)(16000,950.09)(18000,1045.66)(20000,976.34)(22000,1012.98)(24000,978.66)(26000,985.63)(28000,965.61)(30000,934.38)(32000,960.92)(34000,936.29)(36000,973.54)(38000,985.17)(40000,1076.25)};
            \addlegendentry{MAPLE-r}
            
            \addplot[name path=M_low, red_maple, opacity=0, forget plot] coordinates {(0,0.00)(2000,0.00)(4000,96.03)(6000,0.00)(8000,627.38)(10000,903.05)(12000,953.00)(14000,972.17)(16000,1071.04)(18000,1068.60)(20000,1106.71)(22000,1134.39)(24000,1085.87)(26000,1151.17)(28000,1180.24)(30000,1109.30)(32000,1377.22)(34000,1207.95)(36000,1246.77)(38000,1287.58)(40000,1243.70)};
            \addplot[name path=M_high, red_maple, opacity=0, forget plot] coordinates {(0,0.00)(2000,0.00)(4000,282.98)(6000,183.67)(8000,681.55)(10000,939.17)(12000,988.34)(14000,1007.39)(16000,1107.41)(18000,1104.91)(20000,1144.24)(22000,1173.11)(24000,1122.67)(26000,1190.73)(28000,1221.51)(30000,1146.94)(32000,1439.82)(34000,1251.14)(36000,1293.18)(38000,1338.10)(40000,1289.83)};
            \addplot[red_maple, opacity=0.2, forget plot] fill between[of=M_low and M_high];
            \addplot[color=red_maple, mark=none, solid, smooth, tension=0.3, line width=1.1pt] coordinates {(0,0.00)(2000,0.00)(4000,213.07)(6000,91.73)(8000,656.06)(10000,921.32)(12000,970.75)(14000,989.81)(16000,1088.99)(18000,1086.52)(20000,1125.12)(22000,1153.30)(24000,1103.99)(26000,1170.44)(28000,1200.24)(30000,1127.76)(32000,1406.20)(34000,1228.77)(36000,1268.97)(38000,1311.53)(40000,1265.78)};
            \addlegendentry{MAPLE}
            
            \end{axis}
        \end{tikzpicture}
        \caption{11$\times$11 Dark Hex}
        \label{fig:darkhex_elo}
        \end{subfigure}

    \caption{Training Elo ratings of AlphaZe**, MAPLE-r, and MAPLE in two imperfect-information games.
    The shaded regions indicate the 95\% confidence intervals.}
    \label{fig:main_elo}
\end{figure}

The resulting Elo ratings during training for 9$\times$9 Phantom Go and 11$\times$11 Dark Hex are shown in Fig.~\ref{fig:phantomgo_elo} and~\ref{fig:darkhex_elo}, respectively.
For Phantom Go, both MAPLE and MAPLE-r significantly outperform the baseline program AlphaZe**.
At the end of training, AlphaZe**, MAPLE-r, and MAPLE achieve Elo ratings of 1,063, 1,209, and 1,354, corresponding to win rates of 58.96\%, 76.90\%, and 88.45\% against the same opponent pool.
Interestingly, training converges quickly in Phantom Go, with all models plateauing around 20k training steps.
Nevertheless, AlphaZe** remains consistently weaker than the MAPLE-based methods even after 40k training steps.
This is likely because Phantom Go inherits the strategic complexity of Go, making the strategy fusion issue more severe for PIMC-based search.
In contrast, MAPLE-based methods mitigate this issue by aggregating evaluations within a single search tree.

For Dark Hex, MAPLE significantly outperforms both AlphaZe** and MAPLE-r.
At the end of training, AlphaZe**, MAPLE-r, and MAPLE achieve Elo ratings of 1,130, 1,076, and 1,266, corresponding to win rates of 67.87\%, 60.80\%, and 82.20\% against the same opponent pool.
Unlike Phantom Go, MAPLE-r performs slightly worse than AlphaZe**.
Since Dark Hex uses a larger board size, it results in substantially larger information sets and makes the selection of world states more important.
The Siamese-based sampling therefore improves the quality of sampled world states.

Overall, the results demonstrate that MAPLE significantly improves the playing strength of AlphaZero-style agents.
By aggregating evaluations from multiple sampled world states within a single search tree, MAPLE mitigates the strategy fusion problem in PIMC-based search.
Furthermore, the Siamese-based sampling strategy provides additional benefits in environments with larger information sets.

\subsection{Impact of Sampling Size}
To better understand the behavior of MAPLE, this subsection analyzes the impact of sampling size in the information set sampling stage.
Specifically, we examine two parameters: the number of sampled world states $k$ used in the MAPLE tree search and the number of candidate world states $m$ used in the Siamese-based sampling strategy.
For this analysis, we use the final MAPLE checkpoint and evaluate different sampling configurations against the same opponent pool, using the same settings as described in Section~\ref{exp:performance}.

\subsubsection{Number of Sampled World States}
The number of sampled world states $k$ determines how many possible perfect-information environments are aggregated during the MAPLE tree search.
A larger $k$ increases the diversity of sampled world states, but also increases the number of network evaluations, since the total evaluations are $k \times N$, where $N$ is the number of simulations.
To study this effect, we evaluate $k \in \{1,5,10\}$ for both training ($k_\text{T}$) and evaluation ($k_E$).

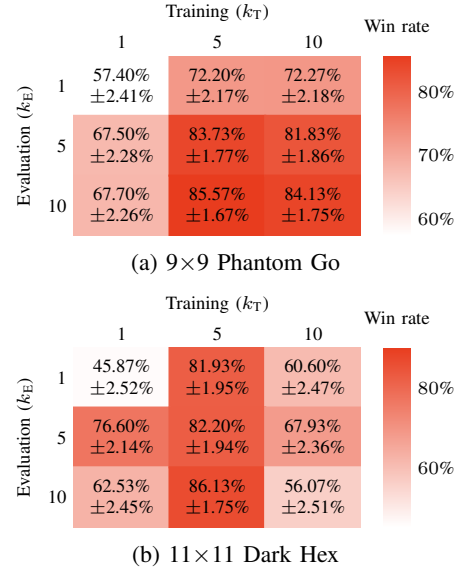
\begin{figure}[tb]
    \centering
    \footnotesize
    \definecolor{myDeepRed}{RGB}{232,60,30}
    \begin{subfigure}[b]{\linewidth}
        \centering
        \begin{tikzpicture}
        \begin{axis}[
            x=36pt,
            y=22.5pt,
            xmin=0.5, xmax=3.5,
            ymin=0.5, ymax=3.5,
            xtick={1,2,3},
            xticklabels={1,5,10},
            ytick={1,2,3},
            yticklabels={1,5,10},
            tick label style={font=\scriptsize},
            tick style={draw=none},
            xtick pos=top,
            xticklabel pos=top,
            y dir=reverse,
            label style={font=\scriptsize},
            xlabel={Training ($k_{\text{T}}$)},
            ylabel={Evaluation ($k_{\text{E}}$)},
            axis on top,
            axis line style={draw=none},
            enlargelimits=false,
            point meta min=57.4,
            point meta max=85.57,
            colormap={myCustomMap}{
                color=(white) 
                color=(myDeepRed)
            },
            colorbar,
            colorbar style={
                width=0.04\linewidth,
                title={Win rate},
                tick label style={font=\scriptsize},
                yticklabel={\pgfmathprintnumber{\tick}\%},
                ytick style={draw=none},
                title style={font=\scriptsize},
                axis line style={draw=none},
            },
            every axis plot/.append style={
                draw=white,
                line width=0.8pt,
            },
        ]
            \addplot [
                matrix plot*,
                mesh/cols=3,
                point meta=explicit,
            ] coordinates {
                (1,1) [57.40]
                (2,1) [72.20]
                (3,1) [72.27]
                (1,2) [67.50]
                (2,2) [83.73]
                (3,2) [81.83]
                (1,3) [67.70]
                (2,3) [85.57]
                (3,3) [84.13]
            };
    
            \newcommand{\celllabel}[4]{
                \node[font=\scriptsize, text=black, align=center] at (axis cs:#1,#2) {#3\% \\ $\pm$#4\%};
            }
    
            \celllabel{1}{1}{57.40}{2.41}
            \celllabel{2}{1}{72.20}{2.17}
            \celllabel{3}{1}{72.27}{2.18}
    
            \celllabel{1}{2}{67.50}{2.28}
            \celllabel{2}{2}{83.73}{1.77}
            \celllabel{3}{2}{81.83}{1.86}
    
            \celllabel{1}{3}{67.70}{2.26}
            \celllabel{2}{3}{85.57}{1.67}
            \celllabel{3}{3}{84.13}{1.75}
    
        \end{axis}
        \end{tikzpicture}
    \caption{9$\times$9 Phantom Go}
    \label{fig:phantomgo_ablation_k}
    \end{subfigure}

    \vspace{0.5em}

    \begin{subfigure}[b]{\linewidth}
        \centering
        \begin{tikzpicture}
        \begin{axis}[
            x=36pt,
            y=22.5pt,
            xmin=0.5, xmax=3.5,
            ymin=0.5, ymax=3.5,
            xtick={1,2,3},
            xticklabels={1,5,10},
            ytick={1,2,3},
            yticklabels={1,5,10},
            tick label style={font=\scriptsize},
            tick style={draw=none},
            xtick pos=top,
            xticklabel pos=top,
            y dir=reverse,
            label style={font=\scriptsize},
            xlabel={Training ($k_{\text{T}}$)},
            ylabel={Evaluation ($k_{\text{E}}$)},
            axis on top,
            axis line style={draw=none},
            enlargelimits=false,
            point meta min=45.0,
            point meta max=90.0,
            colormap={myCustomMap}{
                color=(white) 
                color=(myDeepRed)
            },
            colorbar,
            colorbar style={
                width=0.04\linewidth,
                title={Win rate},
                tick label style={font=\scriptsize},
                yticklabel={\pgfmathprintnumber{\tick}\%},
                ytick style={draw=none},
                title style={font=\scriptsize},
                axis line style={draw=none},
            },
            every axis plot/.append style={
                draw=white,
                line width=0.8pt,
            },
        ]
            \addplot [
                matrix plot*,
                mesh/cols=3,
                point meta=explicit,
            ] coordinates {
                (1,1) [45.87] (2,1) [81.93] (3,1) [60.60]
                (1,2) [76.60] (2,2) [82.20] (3,2) [67.93]
                (1,3) [62.53] (2,3) [86.13] (3,3) [56.07]
            };
    
            \newcommand{\celllabel}[4]{
                \node[font=\scriptsize, text=black, align=center] at (axis cs:#1,#2) {#3\% \\ $\pm$#4\%};
            }
    
            \celllabel{1}{1}{45.87}{2.52}
            \celllabel{2}{1}{81.93}{1.95}
            \celllabel{3}{1}{60.60}{2.47}

            \celllabel{1}{2}{76.60}{2.14}
            \celllabel{2}{2}{82.20}{1.94}
            \celllabel{3}{2}{67.93}{2.36}
    
            \celllabel{1}{3}{62.53}{2.45}
            \celllabel{2}{3}{86.13}{1.75}
            \celllabel{3}{3}{56.07}{2.51}
    
        \end{axis}
        \end{tikzpicture}
    \caption{11$\times$11 Dark Hex}
    \label{fig:darkhex_ablation_k}
    \end{subfigure}
    
     \caption{Win rates with 95\% confidence intervals under different numbers of sampled world states $k$ used during training and evaluation in MAPLE.}
     \label{fig:ablation_k}
\end{figure}

The results in Fig.~\ref{fig:ablation_k} show that increasing $k_\text{E}$ generally improves performance regardless of $k_\text{T}$, especially when increasing from $k_\text{E}=1$ to $k_\text{E}=5$.
For example, in Phantom Go with $k_\text{T}=5$, the win rate improves from 72.2\% with $k_\text{E}=1$ to 83.73\% with $k_\text{E}=5$.
On the other hand, when fixing $k_\text{E}$, training with $k_\text{T}=5$ generally achieves the best performance.
In Phantom Go, $k_\text{T}=5$ and $k_\text{T}=10$ perform similarly and both outperform $k_\text{T}=1$.
In Dark Hex, however, $k_\text{T}=10$ performs even worse than $k_\text{T}=1$.
Considering both performance and computational cost, we adopt $k=5$ as the default setting for both training and evaluation.

\subsubsection{Number of Candidate World States}
The number of candidate world states $m$ determines how many possible world states are sampled from the information set before selecting the final $k$ world states for the MAPLE tree search.
A larger $m$ provides a richer candidate pool for the Siamese network to select more representative world states, but it also increases the computational cost of the sampling process.
For this analysis, we evaluate $m \in \{10, 30, 50, 100\}$ during both training ($m_\text{T}$) and evaluation ($m_\text{E}$).

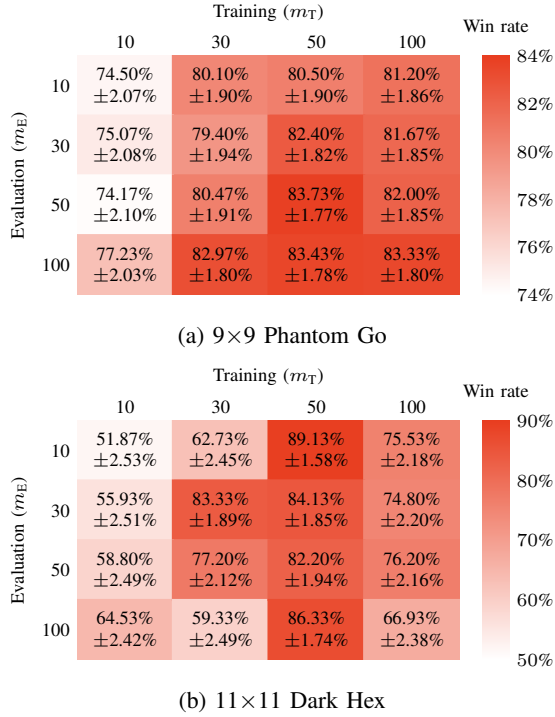
\begin{figure}[tbp]
    \centering
    \footnotesize
    \definecolor{myDeepRed}{RGB}{232,60,30}
    \begin{subfigure}[b]{\linewidth}
        \centering
        \begin{tikzpicture}
        \begin{axis}[
            x=36pt,
            y=22.5pt,
            xmin=0.5, xmax=4.5,
            ymin=0.5, ymax=4.5,
            xtick={1,2,3,4},
            xticklabels={10,30,50,100},
            ytick={1,2,3,4},
            yticklabels={10,30,50,100},
            tick label style={font=\scriptsize},
            tick style={draw=none},
            xtick pos=top,
            xticklabel pos=top,
            y dir=reverse,
            label style={font=\scriptsize},
            xlabel={Training ($m_{\text{T}}$)},
            ylabel={Evaluation ($m_{\text{E}}$)},
            axis on top,
            axis line style={draw=none},
            enlargelimits=false,
            point meta min=74.0,
            point meta max=84.0,
            colormap={myCustomMap}{
                color=(white) 
                color=(myDeepRed)
            },
            colorbar,
            colorbar style={
                width=0.04\linewidth,
                title={Win rate},
                tick label style={font=\scriptsize},
                yticklabel={\pgfmathprintnumber{\tick}\%},
                ytick style={draw=none},
                title style={font=\scriptsize},
                axis line style={draw=none},
            },
            every axis plot/.append style={
                draw=white,
                line width=0.8pt,
            },
        ]
            \addplot [
                matrix plot*,
                mesh/cols=4,
                point meta=explicit,
            ] coordinates {
                (1,1) [74.50] (2,1) [80.10] (3,1) [80.50] (4,1) [81.20]
                (1,2) [75.07] (2,2) [79.40] (3,2) [82.40] (4,2) [81.67]
                (1,3) [74.17] (2,3) [80.47] (3,3) [83.73] (4,3) [82.00]
                (1,4) [77.23] (2,4) [82.97] (3,4) [83.43] (4,4) [83.33]
            };
    
            \newcommand{\celllabel}[4]{
                \node[font=\scriptsize, text=black, align=center] at (axis cs:#1,#2) {#3\% \\ $\pm$#4\%};
            }
    
            \celllabel{1}{1}{74.50}{2.07}
            \celllabel{2}{1}{80.10}{1.90}
            \celllabel{3}{1}{80.50}{1.90}
            \celllabel{4}{1}{81.20}{1.86}
    
            \celllabel{1}{2}{75.07}{2.08}
            \celllabel{2}{2}{79.40}{1.94}
            \celllabel{3}{2}{82.40}{1.82}
            \celllabel{4}{2}{81.67}{1.85}
    
            \celllabel{1}{3}{74.17}{2.10}
            \celllabel{2}{3}{80.47}{1.91}
            \celllabel{3}{3}{83.73}{1.77}
            \celllabel{4}{3}{82.00}{1.85}
    
            \celllabel{1}{4}{77.23}{2.03}
            \celllabel{2}{4}{82.97}{1.80}
            \celllabel{3}{4}{83.43}{1.78}
            \celllabel{4}{4}{83.33}{1.80}
    
        \end{axis}
        \end{tikzpicture}
    \caption{9$\times$9 Phantom Go}
    \label{fig:phantomgo_ablation_m}
    \end{subfigure}

    \vspace{0.5em}

    \begin{subfigure}[b]{\linewidth}
        \centering
        \begin{tikzpicture}
        \begin{axis}[
            x=36pt,
            y=22.5pt,
            xmin=0.5, xmax=4.5,
            ymin=0.5, ymax=4.5,
            xtick={1,2,3,4},
            xticklabels={10,30,50,100},
            xtick pos=top,
            xticklabel pos=top,
            ytick={1,2,3,4},
            yticklabels={10,30,50,100},
            tick label style={font=\scriptsize},
            tick style={draw=none},
            y dir=reverse,
            label style={font=\scriptsize},
            xlabel={Training ($m_{\text{T}}$)},
            ylabel={Evaluation ($m_{\text{E}}$)},
            axis on top,
            axis line style={draw=none},
            enlargelimits=false,
            point meta min=50.0,
            point meta max=90.0,
            colormap={myCustomMap}{
                color=(white) 
                color=(myDeepRed)
            },
            colorbar,
            colorbar style={
                width=0.04\linewidth,
                title={Win rate},
                tick label style={font=\scriptsize},
                yticklabel={\pgfmathprintnumber{\tick}\%},
                ytick style={draw=none},
                title style={font=\scriptsize},
                axis line style={draw=none},
            },
            every axis plot/.append style={
                draw=white,
                line width=0.8pt,
            },
        ]
            \addplot [
                matrix plot*,
                mesh/cols=4,
                point meta=explicit,
            ] coordinates {
                (1,1) [51.87] (2,1) [62.73] (3,1) [89.13] (4,1) [75.53]
                (1,2) [55.93] (2,2) [83.33] (3,2) [84.13] (4,2) [74.80]
                (1,3) [58.80] (2,3) [77.20] (3,3) [82.20] (4,3) [76.20]
                (1,4) [64.53] (2,4) [59.33] (3,4) [86.33] (4,4) [66.93]
            };
    
            \newcommand{\celllabel}[4]{
                \node[font=\scriptsize, text=black, align=center] at (axis cs:#1,#2) {#3\% \\ $\pm$#4\%};
            }
    
            \celllabel{1}{1}{51.87}{2.53}
            \celllabel{2}{1}{62.73}{2.45}
            \celllabel{3}{1}{89.13}{1.58}
            \celllabel{4}{1}{75.53}{2.18}
    
            \celllabel{1}{2}{55.93}{2.51}
            \celllabel{2}{2}{83.33}{1.89}
            \celllabel{3}{2}{84.13}{1.85}
            \celllabel{4}{2}{74.80}{2.20}
    
            \celllabel{1}{3}{58.80}{2.49}
            \celllabel{2}{3}{77.20}{2.12}
            \celllabel{3}{3}{82.20}{1.94}
            \celllabel{4}{3}{76.20}{2.16}
    
            \celllabel{1}{4}{64.53}{2.42}
            \celllabel{2}{4}{59.33}{2.49}
            \celllabel{3}{4}{86.33}{1.74}
            \celllabel{4}{4}{66.93}{2.38}
    
        \end{axis}
        \end{tikzpicture}
    \caption{11$\times$11 Dark Hex}
    \label{fig:darkhex_ablation_m}
    \end{subfigure}
    
    \caption{Win rates with 95\% confidence intervals under different numbers of candidate world states $m$ used during training and evaluation in Siamese-based sampling.}
    \label{fig:ablation_m}
\end{figure}

The results in Fig.~\ref{fig:ablation_m} show that in Phantom Go, increasing $m_\text{T}$ from 10 to 30 leads to clear improvement, while the performance with $m_\text{T}=30, 50, 100$ is similar.
In Dark Hex, the performance improves gradually from $m_\text{T}=10$ to $m_\text{T}=30$, while $m_\text{T}=50$ performs the best.
On the other hand, once a model is trained with a fixed $m_\text{T}$, changing $m_\text{E}$ has little impact on win rate.
This is likely because using a small $m$ during training provides fewer candidate world states, resulting in less informative training data.
Overall, we use $m=50$ as the default setting in our experiments.

\subsection{Evolution of Siamese Embeddings}
\label{exp:siamese_representations}
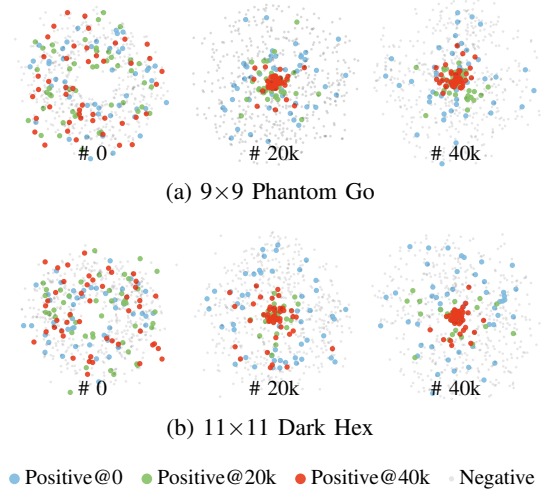
\begin{figure}[tbp]
    \centering

    \definecolor{color1}{RGB}{125,185,222}
    \definecolor{color2}{RGB}{134,193,102}
    \definecolor{color3}{RGB}{232,60,30}

    \begin{tikzpicture}

    \begin{groupplot}[
        group style={
            group size=3 by 2,
            horizontal sep=5pt,
            vertical sep=0.1\linewidth,
            group name=myplots, 
        },
        width=0.95\linewidth,
        height=0.43\linewidth,
        xmin=-3.5, xmax=3.5,
        ymin=-3.5, ymax=3.5,
        axis equal image,
        axis line style={draw=none},
        restrict x to domain=-5:5,
        restrict y to domain=-5:5,
        grid=none,
        title style={
            font=\footnotesize, 
            at={(0.5,0.08)}, 
            anchor=north     
        },
        label style={font=\footnotesize},
        xtick=\empty,
        ytick=\empty,
        legend style={
            draw=none,
            font=\footnotesize,
            /tikz/every even column/.append style={column sep=5pt}
        },
        legend columns=4,
        legend cell align=left,
        legend image post style={fill opacity=1, draw opacity=0, mark size=2pt},
    ]

    \nextgroupplot[title={\# 0}]
    \addplot[only marks, mark=*, fill=color1, draw=none, draw opacity=0, mark size=1pt, mark options={fill opacity=0.9} ] table[x=X, y=Y] {figure/tsne/phantomgo-001-positive-001.txt};
    \addplot[only marks, mark=*, fill=color2, draw=none, draw opacity=0, mark size=1pt, mark options={fill opacity=0.9} ] table[x=X, y=Y] {figure/tsne/phantomgo-001-positive-100.txt};
    \addplot[only marks, mark=*, fill=color3, draw=none, draw opacity=0, mark size=1pt, mark options={fill opacity=0.9} ] table[x=X, y=Y] {figure/tsne/phantomgo-001-positive-200.txt};
    \addplot[only marks, mark=*, fill=black, draw=none, draw opacity=0, mark size=0.5pt, mark options={fill opacity=0.1}, legend image post style={mark size=1pt, fill opacity=0.5} ] table[x=X, y=Y] {figure/tsne/phantomgo-001-negative.txt};

    \nextgroupplot[title={\# 20k}]
    \addplot[only marks, mark=*, fill=color1, draw=none, draw opacity=0, mark size=1pt, mark options={fill opacity=0.9} ] table[x=X, y=Y] {figure/tsne/phantomgo-100-positive-001.txt};
    \addplot[only marks, mark=*, fill=color2, draw=none, draw opacity=0, mark size=1pt, mark options={fill opacity=0.9} ] table[x=X, y=Y] {figure/tsne/phantomgo-100-positive-100.txt};
    \addplot[only marks, mark=*, fill=color3, draw=none, draw opacity=0, mark size=1pt, mark options={fill opacity=0.9} ] table[x=X, y=Y] {figure/tsne/phantomgo-100-positive-200.txt};
    \addplot[only marks, mark=*, fill=black, draw=none, draw opacity=0, mark size=0.5pt, mark options={fill opacity=0.15}, legend image post style={mark size=1pt, fill opacity=0.5} ] table[x=X, y=Y] {figure/tsne/phantomgo-100-negative.txt};

    \nextgroupplot[title={\# 40k}]
    \addplot[only marks, mark=*, fill=color1, draw=none, draw opacity=0, mark size=1pt, mark options={fill opacity=0.9} ] table[x=X, y=Y] {figure/tsne/phantomgo-200-positive-001.txt};
    \addplot[only marks, mark=*, fill=color2, draw=none, draw opacity=0, mark size=1pt, mark options={fill opacity=0.9} ] table[x=X, y=Y] {figure/tsne/phantomgo-200-positive-100.txt};
    \addplot[only marks, mark=*, fill=color3, draw=none, draw opacity=0, mark size=1pt, mark options={fill opacity=0.9} ] table[x=X, y=Y] {figure/tsne/phantomgo-200-positive-200.txt};
    \addplot[only marks, mark=*, fill=black, draw=none, draw opacity=0, mark size=0.5pt, mark options={fill opacity=0.1}, legend image post style={mark size=1pt, fill opacity=0.5} ] table[x=X, y=Y] {figure/tsne/phantomgo-200-negative.txt};

    \nextgroupplot[title={\# 0}, legend to name=scatterlegend]
    \addplot[only marks, mark=*, fill=color1, draw=none, draw opacity=0, mark size=1pt, mark options={fill opacity=0.9} ] table[x=X, y=Y] {figure/tsne/darkhex-001-positive-001.txt};
    \addlegendentry{Positive@0}
    \addplot[only marks, mark=*, fill=color2, draw=none, draw opacity=0, mark size=1pt, mark options={fill opacity=0.9} ] table[x=X, y=Y] {figure/tsne/darkhex-001-positive-100.txt};
    \addlegendentry{Positive@20k}
    \addplot[only marks, mark=*, fill=color3, draw=none, draw opacity=0, mark size=1pt, mark options={fill opacity=0.9} ] table[x=X, y=Y] {figure/tsne/darkhex-001-positive-200.txt};
    \addlegendentry{Positive@40k}
    \addplot[only marks, mark=*, fill=black, draw=none, draw opacity=0, mark size=0.5pt, mark options={fill opacity=0.1}, legend image post style={mark size=1pt, fill opacity=0.5} ] table[x=X, y=Y] {figure/tsne/darkhex-001-negative.txt};
    \addlegendentry{Negative}

    \nextgroupplot[title={\# 20k}]
    \addplot[only marks, mark=*, fill=color1, draw=none, draw opacity=0, mark size=1pt, mark options={fill opacity=0.9} ] table[x=X, y=Y] {figure/tsne/darkhex-100-positive-001.txt};
    \addplot[only marks, mark=*, fill=color2, draw=none, draw opacity=0, mark size=1pt, mark options={fill opacity=0.9} ] table[x=X, y=Y] {figure/tsne/darkhex-100-positive-100.txt};
    \addplot[only marks, mark=*, fill=color3, draw=none, draw opacity=0, mark size=1pt, mark options={fill opacity=0.9} ] table[x=X, y=Y] {figure/tsne/darkhex-100-positive-200.txt};
    \addplot[only marks, mark=*, fill=black, draw=none, draw opacity=0, mark size=0.5pt, mark options={fill opacity=0.1}, legend image post style={mark size=1pt, fill opacity=0.5} ] table[x=X, y=Y] {figure/tsne/darkhex-100-negative.txt};

    \nextgroupplot[title={\# 40k}]
    \addplot[only marks, mark=*, fill=color1, draw=none, draw opacity=0, mark size=1pt, mark options={fill opacity=0.9} ] table[x=X, y=Y] {figure/tsne/darkhex-200-positive-001.txt};
    \addplot[only marks, mark=*, fill=color2, draw=none, draw opacity=0, mark size=1pt, mark options={fill opacity=0.9} ] table[x=X, y=Y] {figure/tsne/darkhex-200-positive-100.txt};
    \addplot[only marks, mark=*, fill=color3, draw=none, draw opacity=0, mark size=1pt, mark options={fill opacity=0.9} ] table[x=X, y=Y] {figure/tsne/darkhex-200-positive-200.txt};
    \addplot[only marks, mark=*, fill=black, draw=none, draw opacity=0, mark size=0.5pt, mark options={fill opacity=0.1}, legend image post style={mark size=1pt, fill opacity=0.5} ] table[x=X, y=Y] {figure/tsne/darkhex-200-negative.txt};

    \end{groupplot}

    \node[anchor=north, yshift=-0.3em, font=\small] at (myplots c2r1.south) {(a) 9$\times$9 Phantom Go \label{fig:phantomgo_tsne}};
    \node[anchor=north, yshift=-9.2em, font=\small] at (myplots c2r1.south) {(b) 11$\times$11 Dark Hex \label{fig:darkhex_tsne}};

    \end{tikzpicture}

    \vspace{0.4em}
    \pgfplotslegendfromname{scatterlegend}

    \caption{Evolutions of Siamese embeddings during training in two games. \# $i$ indicates that the embeddings are generated by the model at the $i$th training step; Positive@$j$ represents the positive samples collected from the self-play at the $j$th step.}
    \label{fig:tsne}
\end{figure}

Since the Siamese network is trained jointly with AlphaZero, its embeddings evolve together with the self-play data generated during training.
This raises the question of how these changes affect sampling quality.
To investigate this, for each game, we extract the self-play datasets and the network models at training steps 0, 20k, and 40k, and cross-evaluate them to form 9 evaluation pairs.
Each dataset consists of 50 states, each with an anchor, a positive, and 50 negative samples.
Then, for each evaluation pair, we generate the embeddings and apply t-SNE \cite{maaten_visualizing_2008} to them, yielding 50 visualizations.
Finally, all the anchors are shifted to the center, and the visualizations are merged into the overall results shown in Fig.~\ref{fig:tsne}, where \#$i$ denotes the model at step $i$, and Positive@$j$ denotes the positive samples from the self-play at step $j$.

Fig.~\ref{fig:tsne} clearly demonstrates that the embeddings of positive samples move closer to the anchor as training proceeds.
For models \#0, all positive samples are widely scattered, reflecting that the untrained network produces random embeddings.
For models \#20k and \#40k, the positive samples begin to cluster around the anchor, showing that the models gradually learn how to predict better embeddings.
Besides, Positive@0 remains widely scattered, which is expected as the later-stage model is no longer trained on the lower-quality data collected during the early self-play.

Specifically, in Phantom Go, models \#20k and \#40k behave similarly, with Positive@0, Positive@20k, and Positive@40k predicted to be the worst, intermediate, and best, respectively.
Interestingly, in Dark Hex, model \#20k predicts Positive@20k best, while model \#40k predicts Positive@40k best. 
This indicates that the model continues to adapt to better fit the data at its current training stage.
Overall, this analysis demonstrates that the Siamese network successfully learns representations during training, reducing the need for pretraining and improving the efficiency of MAPLE.

\section{Discussion}
\label{discussion}
In this paper, we propose MAPLE, a tree search method that aggregates multi-state policy and value evaluations within the AlphaZero framework for IIGs.
MAPLE mitigates the strategy fusion issue in PIMC while maintaining a controllable computational cost compared to IS-MCTS.
Experiments on Phantom Go and Dark Hex demonstrate that MAPLE consistently outperforms the PIMC-based AlphaZero baseline, achieving substantial Elo improvements.
We further provide detailed analyses on the impact of the number of sampled world states and the evolution of the Siamese network, offering insights into the design of MAPLE.

Future directions include exploring more advanced information set sampling strategies, particularly in environments with large uncertainty.
For example, generative models~\cite{li_combining_2025} or belief-based approaches~\cite{wang_beliefstate_2015} can be used to provide more representative world states.
In addition, imperfect-information games often require diverse and adaptive strategies.
Integrating MAPLE with population based training methods (PBT)~\cite{jaderberg_population_2017,wu_accelerating_2020a} or policy-space response oracles (PSRO)~\cite{lanctot_unified_2017} could further improve robustness.
Other analyses, such as investigating the skills learned from MAPLE and examining how much MAPLE can mitigate strategy fusion, would also be interesting.
Overall, MAPLE provides a practical foundation for extending AlphaZero-style learning to IIGs and offers several promising directions for future research.

\section*{Acknowledgement}
This research is partially supported by the National Science and Technology Council (NSTC) of the Republic of China (Taiwan) under Grant Number NSTC 113-2221-E-001-009-MY3,
NSTC 114-2634-F-A49-004,
NSTC 114-2221-E-A49-005,
and NSTC 113-2221-E-A49-128.
The authors acknowledge the use of ChatGPT~\cite{openai_chatgpt_2026} for proofreading, limited to grammar, style, and clarity.
The authors also thank the anonymous reviewers for their valuable comments.

\bibliography{reference}

@article{bertram_neural_2024,
  title = {Neural {{Network-Based Information Set Weighting}} for {{Playing Reconnaissance Blind Chess}}},
  author = {Bertram, Timo and F{\"u}rnkranz, Johannes and M{\"u}ller, Martin},
  year = 2024,
  month = feb,
  journal = {IEEE Transactions on Games},
  volume = {16},
  number = {4},
  pages = {960--970}
}

@inproceedings{bitan_combining_2018,
  title = {Combining {{Prediction}} of {{Human Decisions}} with {{ISMCTS}} in {{Imperfect Information Games}}},
  booktitle = {Proceedings of the 17th {{International Conference}} on {{Autonomous Agents}} and {{MultiAgent Systems}}},
  author = {Bitan, Moshe and Kraus, Sarit},
  year = 2018,
  month = jul,
  series = {{{AAMAS}} '18},
  pages = {1874--1876},
  publisher = {{International Foundation for Autonomous Agents and Multiagent Systems}},
  address = {Richland, SC}
}

@article{bluml_alphaze_2023,
  title = {{{AlphaZe}}{$\ast\ast$}: {{AlphaZero-like}} Baselines for Imperfect Information Games Are Surprisingly Strong},
  author = {Bl{\"u}ml, Jannis and Czech, Johannes and Kersting, Kristian},
  year = 2023,
  journal = {Frontiers in Artificial Intelligence},
  volume = {6}
}

@article{bowling_headsup_2015,
  title = {Heads-up Limit Hold'em Poker Is Solved},
  author = {Bowling, Michael and Burch, Neil and Johanson, Michael and Tammelin, Oskari},
  year = 2015,
  month = jan,
  journal = {Science},
  volume = {347},
  number = {6218},
  pages = {145--149},
  publisher = {American Association for the Advancement of Science}
}

@inproceedings{brown_combining_2020,
  title = {Combining {{Deep Reinforcement Learning}} and {{Search}} for {{Imperfect-Information Games}}},
  booktitle = {Advances in {{Neural Information Processing Systems}}},
  author = {Brown, Noam and Bakhtin, Anton and Lerer, Adam and Gong, Qucheng},
  year = 2020,
  volume = {33},
  pages = {17057--17069},
  publisher = {Curran Associates, Inc.}
}

@inproceedings{brown_libratus_2017,
  title = {Libratus: {{The Superhuman AI}} for {{No-Limit Poker}}},
  booktitle = {Proceedings of the {{Twenty-Sixth International Joint Conference}} on {{Artificial Intelligence}}},
  author = {Brown, Noam and Sandholm, Tuomas},
  year = 2017,
  month = aug,
  pages = {5226--5228},
  publisher = {International Joint Conferences on Artificial Intelligence Organization},
  address = {Melbourne, Australia}
}

@inproceedings{cazenave_phantomgo_2006,
  title = {A {{Phantom-Go Program}}},
  booktitle = {Advances in {{Computer Games}}},
  author = {Cazenave, Tristan},
  year = 2006,
  pages = {120--125},
  publisher = {Springer},
  address = {Berlin, Heidelberg}
}

@inproceedings{coulom_efficient_2007,
  title = {Efficient {{Selectivity}} and {{Backup Operators}} in {{Monte-Carlo Tree Search}}},
  booktitle = {Computers and {{Games}}},
  author = {Coulom, R{\'e}mi},
  year = 2007,
  series = {Lecture {{Notes}} in {{Computer Science}}},
  pages = {72--83},
  publisher = {Springer},
  address = {Berlin, Heidelberg}
}

@article{cowling_information_2012,
  title = {Information {{Set Monte Carlo Tree Search}}},
  author = {Cowling, Peter I. and Powley, Edward J. and Whitehouse, Daniel},
  year = 2012,
  month = jun,
  journal = {IEEE Transactions on Computational Intelligence and AI in Games},
  volume = {4},
  number = {2},
  pages = {120--143}
}

@inproceedings{danihelka_policy_2022,
  title = {Policy Improvement by Planning with {{Gumbel}}},
  booktitle = {International {{Conference}} on {{Learning Representations}}},
  author = {Danihelka, Ivo and Guez, Arthur and Schrittwieser, Julian and Silver, David},
  year = 2022,
  month = apr
}

@article{frank_search_1998,
  title = {Search in Games with Incomplete Information: A Case Study Using {{Bridge}} Card Play},
  author = {Frank, Ian and Basin, David},
  year = 1998,
  month = apr,
  journal = {Artificial Intelligence},
  volume = {100},
  number = {1},
  pages = {87--123}
}

@inproceedings{ginsberg_gib_1999,
  title = {{{GIB}}: {{Steps Toward}} an {{Expert-Level Bridge-Playing Program}}},
  booktitle = {Proceedings of the {{Sixteenth International Joint Conference}} on {{Artificial Intelligence}}},
  author = {Ginsberg, Matthew L.},
  year = 1999,
  month = jul,
  series = {{{IJCAI}} '99},
  pages = {584--593},
  publisher = {Morgan Kaufmann Publishers Inc.},
  address = {San Francisco, CA, USA}
}

@article{ginsberg_gib_2001,
  title = {{{GIB}}: {{Imperfect Information}} in a {{Computationally Challenging Game}}},
  author = {Ginsberg, M. L.},
  year = 2001,
  month = jun,
  journal = {Journal of Artificial Intelligence Research},
  volume = {14},
  pages = {303--358},
  copyright = {Copyright (c)}
}

@inproceedings{kocsis_bandit_2006,
  title = {Bandit {{Based Monte-Carlo Planning}}},
  booktitle = {European {{Conference}} on {{Machine Learning}} and {{Principles}} and {{Practice}} of {{Knowledge Discovery}} in {{Databases}}},
  author = {Kocsis, Levente and Szepesv{\'a}ri, Csaba},
  year = 2006,
  month = sep,
  volume = {2006},
  pages = {282--293}
}

@inproceedings{lanctot_monte_2009,
  title = {Monte {{Carlo Sampling}} for {{Regret Minimization}} in {{Extensive Games}}},
  booktitle = {Advances in {{Neural Information Processing Systems}}},
  author = {Lanctot, Marc and Waugh, Kevin and Zinkevich, Martin and Bowling, Michael},
  year = 2009,
  volume = {22},
  publisher = {Curran Associates, Inc.}
}

@inproceedings{li_combining_2025,
  title = {Combining {{Deep Reinforcement Learning}} and {{Search}} with {{Generative Models}} for {{Game-Theoretic Opponent Modeling}}},
  booktitle = {Thirty-{{Fourth International Joint Conference}} on {{Artificial Intelligence}}},
  author = {Li, Zun and Lanctot, Marc and McKee, Kevin R. and Marris, Luke and Gemp, Ian and Hennes, Daniel and Muller, Paul and Larson, Kate and Bachrach, Yoram and Wellman, Michael P.},
  year = 2025,
  month = sep,
  volume = {1},
  pages = {161--169}
}

@article{long_understanding_2010,
  title = {Understanding the {{Success}} of {{Perfect Information Monte Carlo Sampling}} in {{Game Tree Search}}},
  author = {Long, Jeffrey and Sturtevant, Nathan and Buro, Michael and Furtak, Timothy},
  year = 2010,
  month = jul,
  journal = {Proceedings of the AAAI Conference on Artificial Intelligence},
  volume = {24},
  number = {1},
  pages = {134--140}
}

@article{maaten_visualizing_2008,
  title = {Visualizing {{Data}} Using T-{{SNE}}},
  author = {van der Maaten, Laurens and Hinton, Geoffrey},
  year = 2008,
  journal = {Journal of Machine Learning Research},
  volume = {9},
  number = {86},
  pages = {2579--2605}
}

@article{rosin_multiarmed_2011,
  title = {Multi-Armed Bandits with Episode Context},
  author = {Rosin, Christopher D.},
  year = 2011,
  month = mar,
  journal = {Annals of Mathematics and Artificial Intelligence},
  volume = {61},
  number = {3},
  pages = {203--230}
}

@article{silver_general_2018,
  title = {A General Reinforcement Learning Algorithm That Masters Chess, Shogi, and {{Go}} through Self-Play},
  author = {Silver, David and Hubert, Thomas and Schrittwieser, Julian and Antonoglou, Ioannis and Lai, Matthew and Guez, Arthur and Lanctot, Marc and Sifre, Laurent and Kumaran, Dharshan and Graepel, Thore and Lillicrap, Timothy and Simonyan, Karen and Hassabis, Demis},
  year = 2018,
  month = dec,
  journal = {Science},
  volume = {362},
  number = {6419},
  pages = {1140--1144},
  publisher = {American Association for the Advancement of Science}
}

@article{silver_mastering_2017,
  title = {Mastering the Game of {{Go}} without Human Knowledge},
  author = {Silver, David and Schrittwieser, Julian and Simonyan, Karen and Antonoglou, Ioannis and Huang, Aja and Guez, Arthur and Hubert, Thomas and Baker, Lucas and Lai, Matthew and Bolton, Adrian and Chen, Yutian and Lillicrap, Timothy and Hui, Fan and Sifre, Laurent and {van den Driessche}, George and Graepel, Thore and Hassabis, Demis},
  year = 2017,
  month = oct,
  journal = {Nature},
  volume = {550},
  number = {7676},
  pages = {354--359},
  publisher = {Nature Publishing Group},
  copyright = {2017 Macmillan Publishers Limited, part of Springer Nature. All rights reserved.}
}

@inproceedings{wang_beliefstate_2015,
  title = {Belief-State {{Monte-Carlo}} Tree Search for {{Phantom}} Games},
  booktitle = {2015 {{IEEE Conference}} on {{Computational Intelligence}} and {{Games}} ({{CIG}})},
  author = {Wang, Jiao and Zhu, Tan and Li, Hongye and Hsueh, Chu-Hsuan and Wu, I-Chen},
  year = 2015,
  month = aug,
  pages = {267--274}
}

@article{wu_minizero_2025,
  title = {{{MiniZero}}: {{Comparative Analysis}} of {{AlphaZero}} and {{MuZero}} on {{Go}}, {{Othello}}, and {{Atari Games}}},
  author = {Wu, Ti-Rong and Guei, Hung and Peng, Pei-Chiun and Huang, Po-Wei and Wei, Ting Han and Shih, Chung-Chin and Tsai, Yun-Jui},
  year = 2025,
  month = mar,
  journal = {IEEE Transactions on Games},
  volume = {17},
  number = {1},
  pages = {125--137}
}

@inproceedings{zinkevich_regret_2007,
  title = {Regret {{Minimization}} in {{Games}} with {{Incomplete Information}}},
  booktitle = {Advances in {{Neural Information Processing Systems}}},
  author = {Zinkevich, Martin and Johanson, Michael and Bowling, Michael and Piccione, Carmelo},
  year = 2007,
  volume = {20},
  publisher = {Curran Associates, Inc.}
}

@inproceedings{lanctot_unified_2017,
  title = {A {{Unified Game-Theoretic Approach}} to {{Multiagent Reinforcement Learning}}},
  booktitle = {Advances in {{Neural Information Processing Systems}}},
  author = {Lanctot, Marc and Zambaldi, Vinicius and Gruslys, Audrunas and Lazaridou, Angeliki and Tuyls, Karl and Perolat, Julien and Silver, David and Graepel, Thore},
  year = 2017,
  volume = {30},
  publisher = {Curran Associates, Inc.}
}

@misc{jaderberg_population_2017,
  title = {Population {{Based Training}} of {{Neural Networks}}},
  author = {Jaderberg, Max and Dalibard, Valentin and Osindero, Simon and Czarnecki, Wojciech M. and Donahue, Jeff and Razavi, Ali and Vinyals, Oriol and Green, Tim and Dunning, Iain and Simonyan, Karen and Fernando, Chrisantha and Kavukcuoglu, Koray},
  year = 2017,
  month = nov,
  number = {arXiv:1711.09846},
  eprint = {1711.09846},
  primaryclass = {cs},
  publisher = {arXiv},
  archiveprefix = {arXiv}
}

@article{wu_accelerating_2020a,
  title = {Accelerating and {{Improving AlphaZero Using Population Based Training}}},
  author = {Wu, Ti-Rong and Wei, Ting-Han and Wu, I-Chen},
  year = 2020,
  month = apr,
  journal = {Proceedings of the AAAI Conference on Artificial Intelligence},
  volume = {34},
  number = {01},
  pages = {1046--1053},
  copyright = {Copyright (c) 2020 Association for the Advancement of Artificial Intelligence}
}

@inproceedings{buro_improving_2009,
  title = {Improving State Evaluation, Inference, and Search in Trick-Based Card Games},
  booktitle = {Proceedings of the 21st {{International Joint Conference}} on {{Artificial Intelligence}}},
  author = {Buro, Michael and Long, Jeffrey R. and Furtak, Timothy and Sturtevant, Nathan},
  year = 2009,
  series = {{{IJCAI}}'09},
  pages = {1407--1413},
  publisher = {Morgan Kaufmann Publishers Inc.},
  address = {San Francisco, CA, USA}
}

@inproceedings{whitehouse_determinization_2011,
  title = {Determinization and Information Set {{Monte Carlo Tree Search}} for the Card Game {{Dou Di Zhu}}},
  booktitle = {2011 {{IEEE Conference}} on {{Computational Intelligence}} and {{Games}} ({{CIG}}'11)},
  author = {Whitehouse, Daniel and Powley, Edward J. and Cowling, Peter I.},
  year = 2011,
  month = aug,
  pages = {87--94}
}

@misc{openai_chatgpt_2026,
  author       = {OpenAI},
  title        = {Chat{GPT}: Language Model by {OpenAI}},
  year         = {2026},
  howpublished = {\url{https://chat.openai.com/}},
}
\bibliographystyle{IEEEtran}

\end{document}